\documentclass[runningheads]{llncs}
\usepackage{graphicx}

\usepackage{tikz}
\usepackage{comment}
\usepackage{amsmath,amssymb} 
\usepackage{color}

\usepackage[accsupp]{axessibility}  

\usepackage[width=122mm,left=12mm,paperwidth=146mm,height=193mm,top=12mm,paperheight=217mm]{geometry}

\begin{document}
\pagestyle{headings}
\mainmatter
\def\ECCVSubNumber{800}  

\title{Masked Autoencoders for Point Cloud Self-supervised Learning} 

\titlerunning{Point-MAE}
%
\author{Yatian Pang\inst{2} \quad
Wenxiao Wang\inst{3} \quad
Francis E.H. Tay\inst{2} \quad  \\
Wei Liu\inst{4} \quad
Yonghong Tian\inst{5} \quad
Li Yuan\inst{1}\thanks{Corresponding author}}

\authorrunning{Y. Pang et al.}
%
\institute{School of ECE at Peking University, Shenzhen Graduate School \and
National University of Singapore \and
ZheJiang University \and
Tencent Data Platform \and
School of Computer Science at Peking University \& Pengcheng Laboratory \\
\email{yatian\_pang@u.nus.edu; yuanli-ece@pku.edu.cn}}

\maketitle

\begin{abstract}
As a promising scheme of self-supervised learning, masked autoencoding has significantly advanced natural language processing and computer vision. Inspired by this, we propose a neat scheme of masked autoencoders for point cloud self-supervised learning, addressing the challenges posed by point cloud's properties, including leakage of location information and uneven information density. Concretely, we divide the input point cloud into irregular point patches and randomly mask them at a high ratio. Then, a standard Transformer based autoencoder, with an asymmetric design and a shifting mask tokens operation, learns high-level latent features from unmasked point patches, aiming to reconstruct the masked point patches. Extensive experiments show that our approach is efficient during pre-training and generalizes well on various downstream tasks. Specifically, our pre-trained models achieve 85.18\% accuracy on ScanObjectNN and 94.04\% accuracy on ModelNet40, outperforming all the other self-supervised learning methods. We show with our scheme, a simple architecture entirely based on standard Transformers can surpass dedicated Transformer models from supervised learning. Our approach also advances state-of-the-art accuracies by 1.5\%-2.3\% in the few-shot object classification. Furthermore, our work inspires the feasibility of applying unified architectures from languages and images to the point cloud. Codes are available at https://github.com/Pang-Yatian/Point-MAE.
\end{abstract}

\section{Introduction}

\begin{figure}
    \centering
    \includegraphics[width=\textwidth]{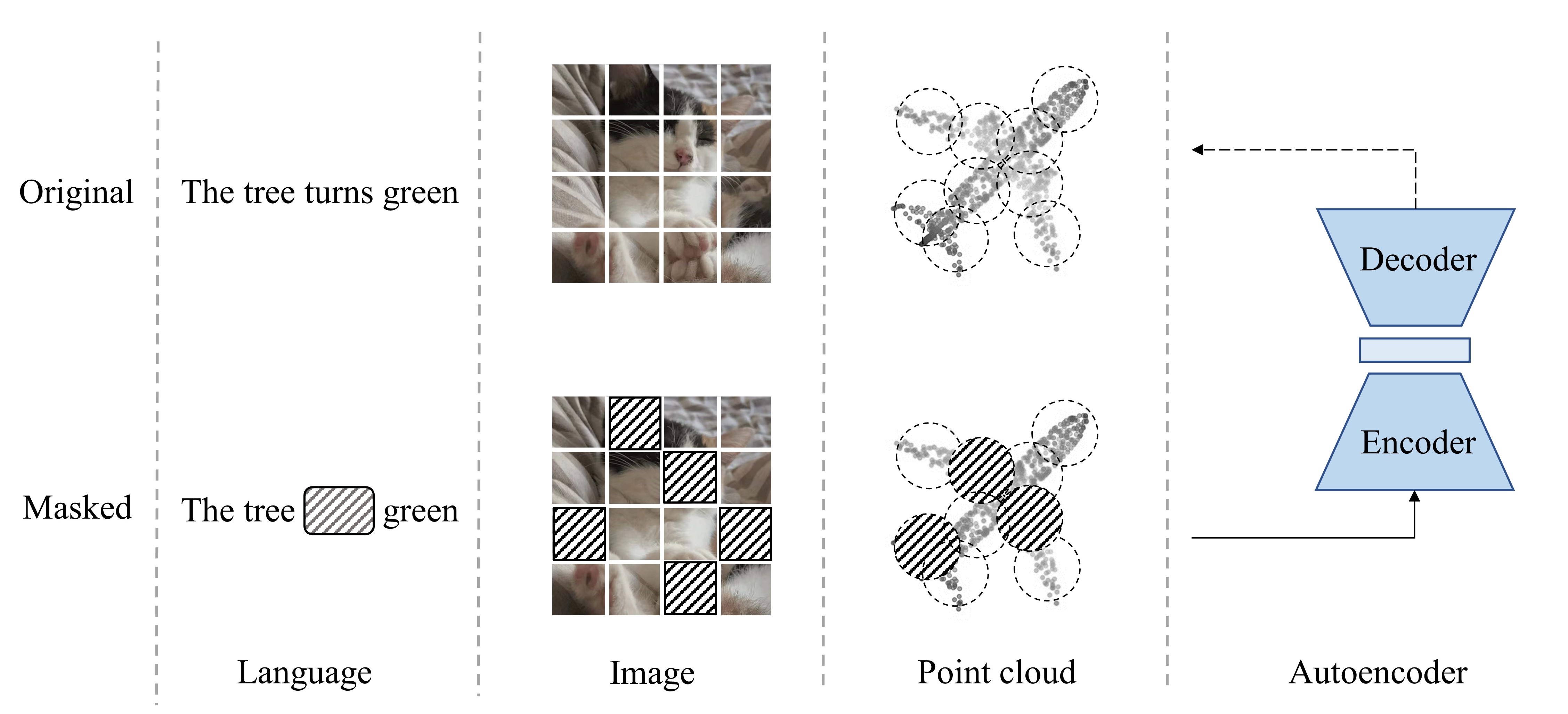}
    \caption{{\bf Illustration of masked autoencoding.} A portion of input data is masked, then an autoencoder is trained to recover the masked parts from original input data. The encoder in autoencoder is encouraged to learn high-level latent features from unmasked parts.}
    \label{fig:auto}
\end{figure}

Self-supervised learning learns latent features from unlabeled data instead of building representations based on human-defined annotations. It is usually done by designing a pretext task to pre-train the model, then fine-tune on downstream tasks. Relying less on labeled data, self-supervised learning has significantly advanced natural language processing (NLP)~\cite{bert,nlp1,nlp3,nlp2} and computer vision~\cite{cv1,cv2,cvcontrastive1,cvcontrastive2,cvgenerative1,data2vec,mae,simmim}. Among them, masked autoencoding~\cite{mae,simmim,data2vec}, illustrated in Figure \ref{fig:auto}, is a promising scheme for both languages and images. It randomly masks a portion of input data and adopts an autoencoder to reconstruct explicit features (e.g., pixels) or implicit features (e.g., discrete tokens) corresponding to the original masked content. As masked parts do not provide data information, this reconstruction task enables the autoencoder to learn high-level latent features from unmasked parts. Besides, the powerful capability of masked autoencoding gives credit to its autoencoder's backbone, which adopts Transformers~\cite{tf} architecture. For example, BERT~\cite{bert} in NLP and MAE~\cite{mae} in computer vision both apply masked autoencoding and adopt a standard Transformer architecture as autoencoder's backbone to achieve state-of-the-art performance.

The idea of masked autoencoding is also applicable for point cloud self-supervised learning, as point cloud essentially shares a common property with both languages and images (see Figure \ref{fig:auto}). Specifically, the fundamental elements (i.e., points, vocabularies, and pixels) that carry information are not independent. Instead, neighbouring elements form a meaningful subset to present local features. Together with local features, the complete set of elements makes up global features. Therefore, after embedding point subsets into tokens, the point cloud can be processed similarly with languages and images. Furthermore, considering datasets for the point cloud are relatively small, masked autoencoding as a self-supervised learning method can naturally address the large data demand of Transformers architecture, which is the autoencoder's backbone. Indeed, a recent work Point-BERT~\cite{pointbert} attempts a scheme somewhat similar to masked autoencoding. It proposes a BERT-style pre-training strategy by masking input tokens of the point cloud, then adopts a Transformer architecture to predict discrete tokens of the masked tokens. However, this method is relatively sophisticated as it is required to train a DGCNN~\cite{ptdgcnn} based discrete Variational AutoEncoder (dVAE)~\cite{dvae} before pre-training and relies heavily on contrastive learning as well as data augmentation during pre-training. Moreover, the masked tokens from their inputs are processed from the input of Transformers during pre-training, leading to early leakage of location information and high consumption of computing resources. Different from their method, and more importantly, to introduce masked autoencoding to the point cloud, we aim to design a neat and efficient scheme of masked autoencoders. To this end, we first analyze the main challenges of introducing masked autoencoding for point cloud from the following aspects:

(i) Lack of a unified Transformer architecture. Compared to Transformers~\cite{tf} in NLP and Vision Transformer (ViT)~\cite{cvvit} in computer vision, Transformer architectures for point cloud are less studied and relatively diverse, mainly because small datasets cannot meet the large data demand of Transformers. Different from previous methods that use dedicated Transformers or adopt extra non-Transformers models to assist (such as Point-BERT~\cite{pointbert} uses an extra DGCNN~\cite{ptdgcnn}), we aim to build our autoencoder's backbone entirely based on standard Transformers, which can serve as a potential unified architecture for point cloud.

(ii) Positional embeddings for mask tokens lead to leakage of location information. In masked autoencoders, each masked part is replaced by a share-weighted learnable mask token. All the mask tokens need to be provided with their location information in input data by positional embeddings. Then after processing by autoencoders, each mask token is used to reconstruct the corresponding masked part. Providing location information is not an issue for languages and images, because they do not contain location information. While point cloud naturally has location information in the data, leakage of location information to mask tokens makes the reconstruction task less challenging, which is harmful for autoencoders learning latent features. We address this issue by shifting mask tokens from the input of the autoencoder's encoder to the input of the autoencoder's decoder. This delays the leakage of location information and enables the encoder to focus on learning features from unmasked parts. 

\begin{figure}
    \centering
    \includegraphics[width = \textwidth]{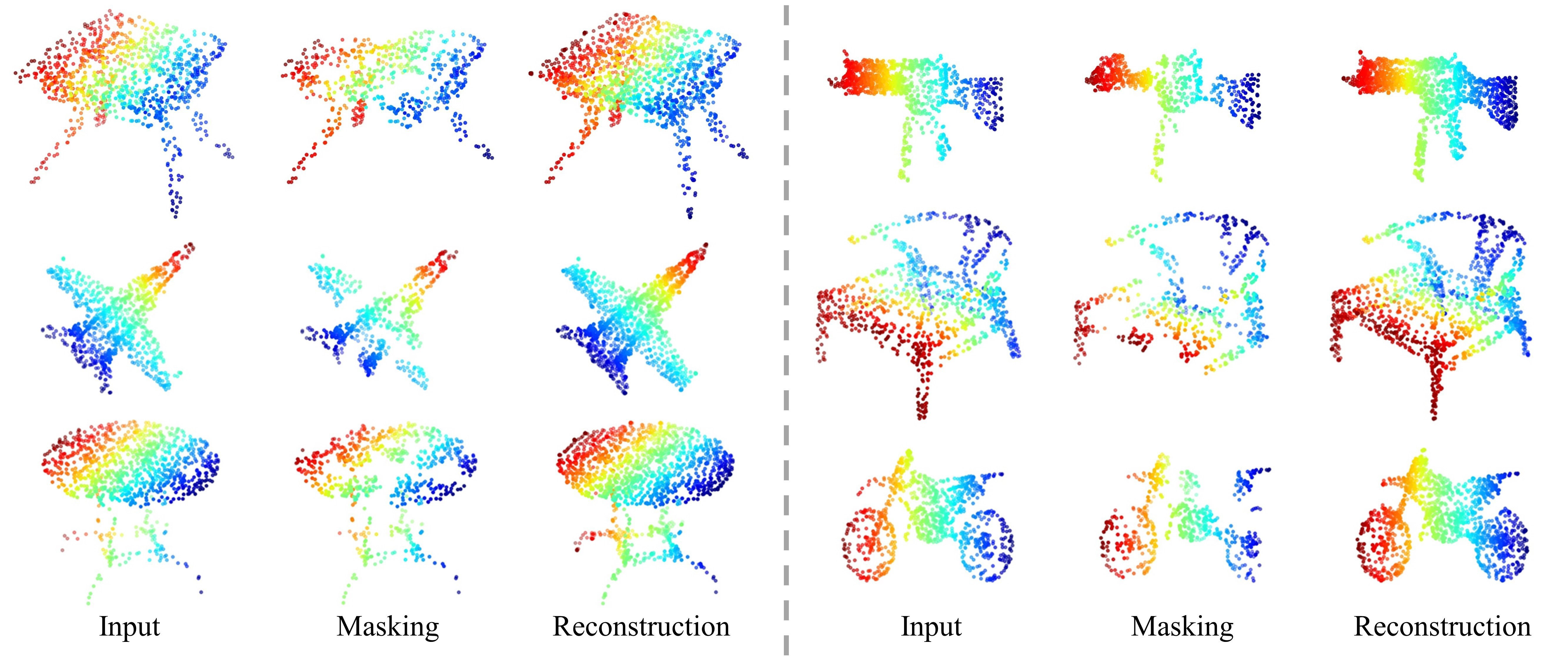}
    \caption{{\bf Reconstruction examples on ShapeNet validation set.} In each group, we show the original input (i.e., ground truth), masked point cloud, and reconstruction result from left to right. The masking ratio is 60\%. It can be observed directly that reconstructions of key local features (such as sharp corners) are much worse than reconstructions of less important local features (such as flat surfaces).}
    \label{fig:vis1}
\end{figure}

(iii) Point cloud carries information in a different density compared to languages and images. Languages contain high-density information, while images contain heavy redundant information~\cite{mae}. In the point cloud, information density distribution is relatively uneven. The points that make up key local features (e.g., sharp corners and edges) contain a much higher density of information than the points that make up less important local features (e.g., flat surfaces). In other words, if being masked, the points that contain high-density information is more difficult to be recovered in the reconstruction task. This can be directly observed in reconstruction examples, as shown in Figure \ref{fig:vis1}. Taking the last row of Figure \ref{fig:vis1} for illustration, the masked desk surface (left) can be easily recovered, while the reconstruction of the masked motorcycle's wheel (right) is much worse. Although the point cloud contains uneven density of information, we find that random masking at a high ratio (60\%-80\%) works well, which is surprisingly the same as images. This indicates the point cloud is similar to images instead of languages, in terms of information density.

Driven by the analysis, we propose a novel self-supervised learning framework for \textbf{Point} cloud by
designing a neat and efficient scheme of \textbf{M}asked \textbf{A}uto\textbf{E}ncoders, termed as \textbf{Point-MAE}. As shown in Figure \ref{fig:net}, our Point-MAE mainly consists of a point cloud masking and embedding module, and an autoencoder. The input point cloud is divided into irregular point patches, which are randomly masked at a high ratio to reduce data redundancy. Then, the autoencoder learns high-level latent features from unmasked point patches, aiming to reconstruct masked point patches in coordinate space. Specifically, our autoencoder's backbone is entirely built by standard Transformer blocks and adopts an asymmetric encoder-decoder structure~\cite{mae}. The encoder only processes unmasked point patches. Then taking both encoded tokens and mask tokens as input, the lightweight decoder with a simple prediction head reconstructs masked point patches. Compared to processing mask tokens from the input of the encoder, shifting mask tokens to the lightweight decoder results in significant computational savings, and more importantly, avoiding early leakage of location information.

Our approach is effective, and pre-trained models generalize well on various downstream tasks. In object classification tasks, our Point-MAE achieves 85.18\% accuracy on the hardest setting of real-world dataset ScanObjectNN and 94.04\% accuracy on a clean object dataset ModelNet40, outperforming all the other self-supervised learning methods. Meanwhile, Point-MAE surpasses all the dedicated Transformers models from supervised learning. In the few-shot object classification, Point-MAE significantly advances state-of-the-art accuracies by 1.5\%-2.3\% on different settings of ModelNet40. When generalized to the part segmentation task, Point-MAE largely improves the baseline by 1\% mean IoU.

Our main contributions can be summarized as follows: 

(1) We propose a novel scheme of masked autoencoders for point cloud self-supervised learning, addressing key issues including backbone architecture, early leakage of location information, and information density of the point cloud. Our approach is neat and efficient, with high generalization capability on various downstream tasks, outperforming all the other self-supervised learning methods. 

(2) We show with our approach, a simple architecture that is entirely based on standard Transformers can surpass dedicated Transformer models from supervised learning. This result suggests that standard Transformers can serve as a potential unified architecture in the point cloud discipline.  

(3) From the perspective of multimodal learning, our work inspires that unified architectures for languages and especially images, such as masked autoencoders, are also applicable for point cloud, when equipped with a modality-specific embedding module and a task-specific output head. We hope our field could be further advanced with the joint of other modality data.

\section{Related Work}


\subsection{Self-supervised Learning}
In the machine learning field, Self-supervised Learning (SSL) is defined as "the machine predicts any parts of its input for any observed part"\footnote{https://aaai.org/Conferences/AAAI-20/invited-speakers/}. The main ideas can be summarized as: a) supervision labels are generated from the data itself instead of human annotating, b) the model predicts parts of the data from other parts~\cite{ssl}. This process is usually done by designing a pretext task, which relieves the high demand for manual labeling data.

\subsubsection{SSL for NLP and Image}

In the NLP field, SSL has been well developed. Generative SSL methods such as BERT~\cite{bert} gain huge success by designing pretext tasks that mask input tokens, and pre-train the model to predict original vocabularies. In computer vision for images, contrastive SSL methods~\cite{cvcontrastive1,cvcontrastive2,cvcontrastive3,cvcontrastive4,cvcontrastive5} aim to discriminate the degree of similarities between different augmented images. These methods have dominated until recent generative SSL methods~\cite{mae,simmim,cvgenerative2} result in more competitive performance. For example, MAE~\cite{mae} randomly masks input patches, and pre-train the model to recover masked patches in pixel space.

\subsubsection{SSL for Point Cloud}

SSL has also been widely studied for point cloud representation learning~\cite{ptocco,ptdeepcon,ptssl1,ptssl2,ptssl3,ptssl4,foldingnet,ptssl6,ptssl7,ptiae}. Pretext tasks are relatively diverse. Among them, DepthContrast~\cite{ptdeepcon} sets an instance discrimination task for two augmented versions of an input point cloud. OcCo~\cite{ptocco} attempts to recover the original point cloud from the occluded point cloud in camera views. IAE~\cite{ptiae} adopts an autoencoder to reconstruct implicit features from augmented inputs. A recent work Point-BERT~\cite{pointbert} proposes a BERT-style pre-training strategy by masking input tokens and aims to predict discrete tokens of masked parts, with the assistance of dVAE~\cite{dvae}. Different from previous methods, we attempt to design a neat scheme for point cloud self-supervised learning.

\subsection{Autoencoders}

Generally, an autoencoder consists of an encoder followed by a decoder. The encoder is responsible for encoding inputs to high-level latent features. Then the decoder decodes latent features, aiming to reconstruct the input. The optimization goal is to make the reconstructed data as similar as possible to the original input, such as mean squared error loss in pixel space for images. 

Specifically, our approach belongs to the class of denoising autoencoders. The main idea of denoising autoencoders is to enhance the robustness of the model by introducing input noise. Following the same principle, masked autoencoders introduce input noise through a masking operation. For example, in NLP, BERT~\cite{bert} adopts masked language modeling. It randomly masks tokens from the input, then applies an autoencoder to predict vocabularies corresponding to masked tokens. In computer vision, both MAE~\cite{mae} and SimMIM~\cite{simmim} propose a similar masked image modeling, which randomly masks input image patches. Then autoencoders are applied to predict the masked patches in pixel space. Inspired by the above ideas, our work aim to introduce masked autoencoders to point cloud.

\subsection{Transformers}
Transformers~\cite{tf} model global dependencies of input through the self-attention mechanism, and have dominated in NLP~\cite{bert,nlp1,nlp2,nlp3,nlp4}. Since ViT~\cite{cvvit}, Transformers architectures have been popular in computer vision~\cite{cvt2t,tfswin,cvvolo,cvcross,cvpvt,ptpt,ptpct,chen2021full}. However, as backbones for masked autoencoders, Transformers architectures for point cloud representation learning are less developed. PCT~\cite{ptpct} designs a dedicated input embedding layer and modifies the self-attention mechanism in Transformer layers. PointTransformer~\cite{ptpt} also modifies the Transformer layer, and uses extra aggregating operations between Transformer blocks. The recent work Point-BERT~\cite{pointbert} introduces a standard Transformer architecture, but requires DGCNN~\cite{ptdgcnn} to assist pre-training. Different from previous works, our work presents an architecture that is entirely based on standard Transformers.

\begin{figure}
    \centering
    \includegraphics[width=\textwidth]{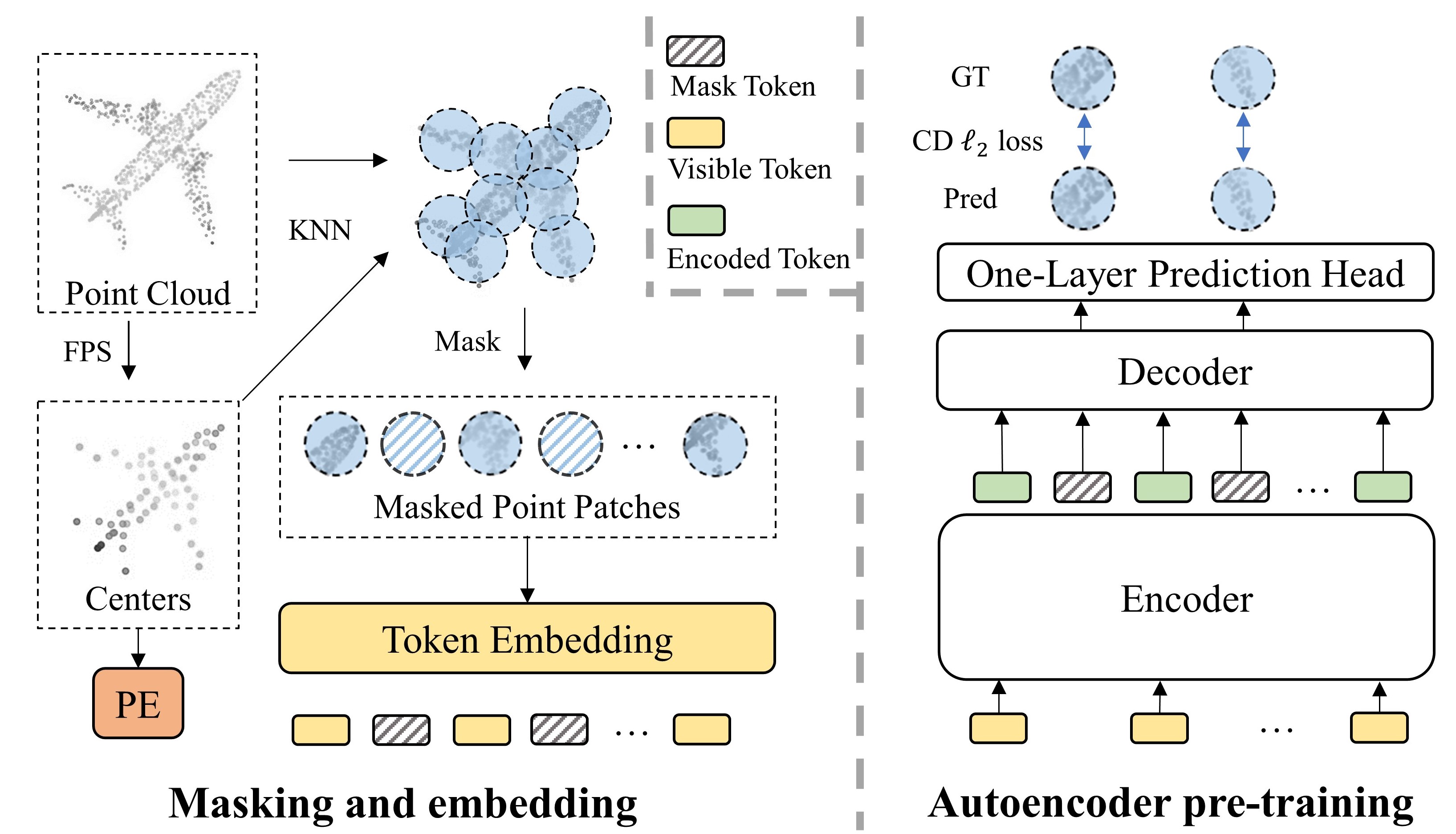}
    \caption{{\bf Overall scheme of our Point-MAE.} On the left, we show the masking and embedding process. The input cloud is divided into point patches, which are masked randomly and then embedded. Autoencoder pre-training is shown on the right. The encoder only processes visible tokens. Mask tokens are added to the input sequence of the decoder to reconstruct masked point patches.}
    \label{fig:net}
\end{figure}

\section{Point-MAE}
We aim to design a neat and efficient scheme of masked autoencoders for point cloud self-supervised learning. Figure \ref{fig:net} illustrates the overall scheme of our approach Point-MAE. The input point cloud is first processed by a masking and embedding module. Then a standard Transformer based autoencoder is adopted, including a simple prediction head, to reconstruct the masked parts of the input point cloud.

\subsection{Point Cloud Masking and Embedding}

Unlike images in computer vision that can be naturally divided into regular patches, point cloud consists of unordered points in 3D space. Based on its property, we process the input point cloud through three stages: point patches generation, masking, and embedding.

\subsubsection{Point Patches Generation}
Following Point-BERT~\cite{pointbert}, we divide input point cloud into irregular point patches (may overlap) via Farthest Point Sampling (FPS) and K-Nearest Neighborhood (KNN) algorithm. Formally, given an input point cloud with \(p\) points \(X^i \in \mathbb{R}^{p\times3}\), FPS is applied to sample \(n\) points for centers \(CT\) in point patches. Based on center points, KNN selects \(k\) nearest points from input for corresponding point patches \(P\),
\begin{align}
    CT = FPS(X^i),& \quad CT \in \mathbb{R}^{n\times3}; \\
    \label{eq:patches}
    P = KNN(X^i, CT),& \quad P \in \mathbb{R}^{n\times k\times3}.
\end{align}
   
Note that in point patches, each point is represented by normalized coordinates with respect to its center point. This leads to better convergence.

\subsubsection{Masking}

Considering point patches may overlap, we mask them separately, in order to keep information complete in each point patch. With a masking ratio \(m\), the set of masked patches is denoted as \(P_{gt} \in \mathbb{R}^{mn\times k\times 3}\), which is used as ground truth in the computing of reconstruction loss. As for masking strategy, we find random masking at a high ratio (60\%-80\%) works well for our approach, see Section \ref{sec:mask}.


\subsubsection{Embedding}
For the embedding of each masked point patch, we replace it with a share-weighted learnable mask token. We denote the full set of mask tokens as \(T_m \in \mathbb{R}^{mn\times C}\), where \(C\) is the embedding dimension. For the unmasked (visible) point patches, a naive idea is to flatten and embed them with a trainable linear projection, similar to ViT~\cite{cvvit}. However, we argue that linear embedding fails to follow the principle of permutation invariance~\cite{pointnet}. A more reasonable embedding method should be adopted. To keep neat, we implement a lightweight PointNet~\cite{pointnet}, which mainly consists of MLPs and max pooling layers. The visible point patches \(P_{v} \in \mathbb{R}^{(1-m)n\times k\times 3}\) are hence embedded into visible tokens \(T_v\),
\begin{equation}
    T_v = PointNet(P_{v}), \quad T_v \in \mathbb{R}^{(1-m)n\times C}.
\end{equation}

Considering point patches are represented in normalized coordinates, providing centers' position information to embedding tokens is essential. A simple method for Position Embedding (PE) is mapping coordinates of centers to embedding dimension with a learnable MLP, following previous works~\cite{pointbert,ptpt}. Note that we use two separate PE for encoder and decoder respectively in our autoencoder, introduced next.
\subsection{Autoencoder's Backbone}

Our autoencoder's backbone is entirely based on standard Transformers, with an asymmetric encoder-decoder design~\cite{mae}. The last layer of the autoencoder adopts a simple prediction head to achieve the reconstruction target.

\subsubsection{Encoder-decoder}

 Our encoder consists of standard Transformer blocks and only encodes visible tokens \(T_v\) without mask tokens \(T_m\). The encoded tokens are denoted as \(T_e\). Furthermore, positional embeddings are added to every Transformer block, providing location information. 
 
Our decoder is similar to the encoder but contains fewer Transformer blocks. It takes both encoded tokens \(T_e\) and masks tokens \(T_m\) as input. A full set of positional embeddings is added to every Transformer block, providing location information to all the tokens. After processing, the decoder only outputs the decoded mask tokens \(H_m\), which are fed to the following prediction head. The encoder-decoder structure is formulated as,
 \begin{align}
        T_e = Encoder(T_v&), \quad T_e \in \mathbb{R}^{(1-m)n\times C} ; \\
         H_m = Decoder(concat(&T_e, T_m)), \quad H_m \in \mathbb{R}^{mn\times C} .
 \end{align}
 
 In our encoder-decoder structure, we shift the mask tokens to the lightweight decoder instead of processing them from the input of the encoder. This design is beneficial from two aspects. First, as we use high masking ratios, shifting mask tokens significantly reduces the number of input tokens for the encoder. Therefore, we can save computational resources due to the quadratic complexity of Transformers. More importantly, shifting mask tokens to the decoder can avoid early leakage of location information to the encoder, making the encoder learn latent features better (see Section \ref{sec:mask}).

\subsubsection{Prediction Head}
As the last layer of backbone, the prediction head aims to reconstruct masked point patches in coordinate space. We simply use a fully connected (FC) layer as our prediction head. Taking the output \(H_m\) from the decoder, the prediction head projects it to a vector, which has the same number of dimensions as the total number of coordinates in a point patch. Then followed by a reshape operation, predicted masked point patches \(P_{pre}\) are obtained,

 
\begin{align}
    P_{pre} = Reshape(FC(H_m)),  \quad P_{pre} \in \mathbb{R}^{mn\times k\times 3}.
\end{align}

\subsection{Reconstruction Target}
Our reconstruction target is to recover coordinates of the points in every masked point patch. Given the predicted point patches \(P_{pre}\) and ground truth \(P_{gt}\), we compute the reconstruction loss by \(l_2\) Chamfer Distance~\cite{cd},
\begin{equation}
      L = \frac{1}{|P_{pre}|} \sum_{a\in P_{pre}} \mathop{\min}\limits_{b \in P_{gt}} \left \| a-b \right \|_2^2
       + \frac{1}{|P_{gt}|} \sum_{b\in P_{gt}} \mathop{\min}\limits_{a \in P_{pre}} \left \| a-b \right \|_2^2
\end{equation}

\section{Experiments}
We conduct the following experiments with our Point-MAE. a) We pre-train our model on ShapeNet~\cite{shapenet} training set. b) We evaluate our pre-trained model on various downstream tasks, including object classification, few-shot learning and part segmentation. c) We study different masking strategies, and we show the effect of shifting mask tokens.

In our Point-MAE, for different resolutions of the input point cloud, we divide them into different numbers of patches with a linear scaling. A typical input with \(p = 1024\) points is divided into \(n = 64\) point patches. For the KNN algorithm, we set \(k = 32\) to keep the number of points in each patch constant. In the autoencoder's backbone, the encoder has 12 Transformer blocks while the decoder has 4 Transformer blocks. Each Transformer block has 384 hidden dimensions and 6 heads. MLP ratio in Transformer blocks is set to 4.

\subsection{Pre-training Setup}

ShapeNet~\cite{shapenet} consists of about 51,300 clean 3D models, covering 55 common object categories. We split the dataset into a training set and a validation set but only conduct pre-training on the training set. For each instance, we sample 1024 points via FPS as input point cloud. Note that we only apply standard random scaling and random translation for data augmentation during pre-training. For pre-training details, we use an AdamW optimizer~\cite{adamw} and cosine learning rate decay~\cite{cos}. The initial learning rate is set to 0.001, with a weight decay of 0.05. We pre-train our model for 300 epochs, with a batch size of 128.

\begin{figure}
    \centering
    \includegraphics[width=\textwidth]{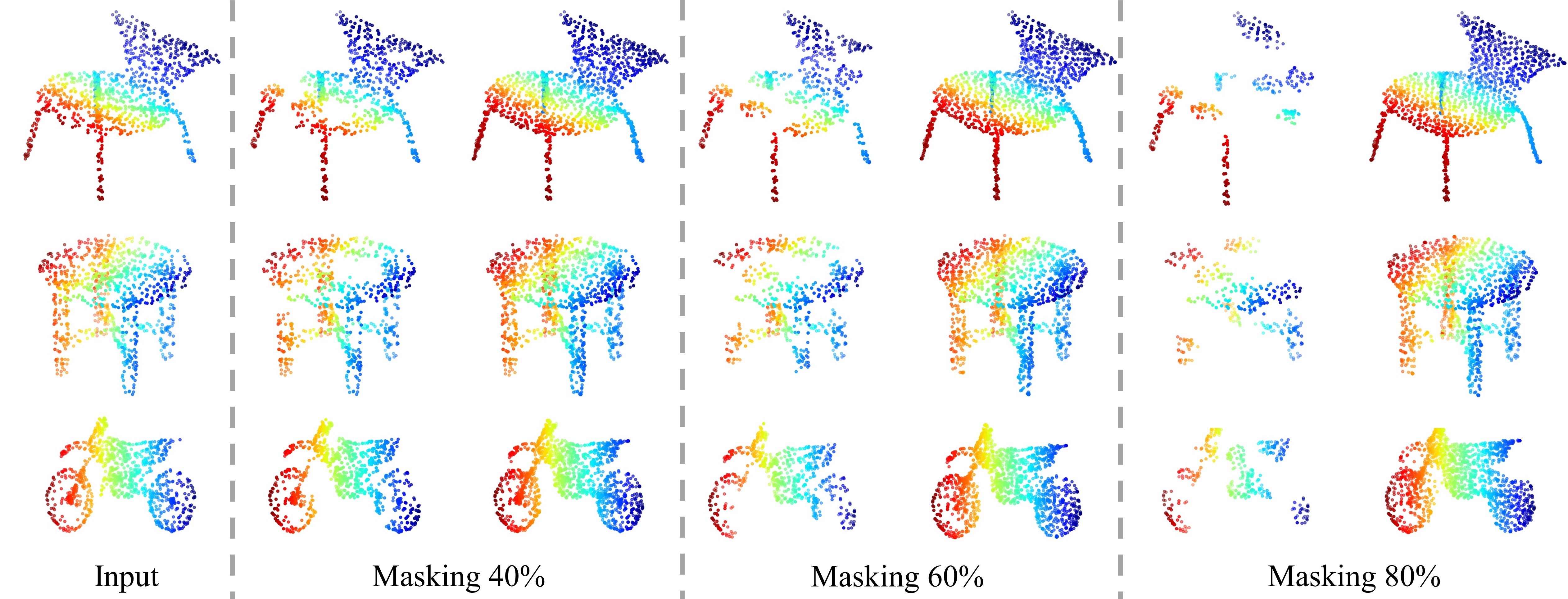}
    \caption{{\bf Reconstruction results on ShapeNet validation set.}  The model is pre-trained with a masking ratio of 60\% but can generalize well on inputs with different masking ratios. Inputs are shown in the leftmost column. In the following columns, we show the masked input (left) and reconstruction (right) with different masking ratios.}
    \label{fig:pre_vis}
    \vspace{-5mm}
\end{figure}

To demonstrate the effectiveness of our method, we visualize reconstruction results on ShapeNet validation set in Figure \ref{fig:pre_vis}. The model is pre-trained with a masking ratio of 60\%, but it is able to reconstruct inputs with different masking ratios. This high generalization capability can be expected, as our model learns high-level latent features well. Furthermore, our method speeds up pre-training by 1.7$\times$ compared to Point-BERT~\cite{pointbert}.

\subsection{Downstream Tasks}

\subsubsection{Object Classification on Real-World Dataset}

In SSL for point cloud, one of the main concerns is to design a model with high generalization capability. Specifically, the commonly used dataset for pre-training, ShapeNet~\cite{shapenet}, only contains clean object models, without any scene context such as backgrounds. Motivated by this, we evaluate our pre-trained model on a challenging real-world dataset, ScanObjectNN~\cite{scan}, which consists of about 15,000 objects from 15 categories. The objects are scanned from real-world indoor scene data with cluttered backgrounds.

\begin{table}
\begin{center}
\caption{{\bf Object classification on real-world ScanObjectNN dataset.} We evaluate our approach on three variants, among which PB-T50-RS is the hardest setting. Accuracy (\%) for each variant is reported.}
\setlength\tabcolsep{5pt}
\label{tb:2}
\begin{tabular}{l|ccc}
\hline
Methods & OBJ-BG & OBJ-ONLY & PB-T50-RS  \\

\hline

 PointNet~\cite{pointnet}& 73.3 & 79.2 & 68.0 \\
 SpiderCNN~\cite{ptspidercnn}& 77.1 & 79.5 & 73.7 \\
 PointNet\(++\)~\cite{pointnet++}& 82.3 & 84.3 & 77.9 \\
 DGCNN~\cite{ptdgcnn}& 82.8& 86.2 & 78.1 \\
 PointCNN~\cite{ptptcnn}& 86.1 & 85.5 & 78.5 \\
 BGA-DGCNN~\cite{scan}& - & - & 79.7 \\
 BGA-PN\(++\)~\cite{scan}& - & - & 80.2 \\
 GBNet~\cite{ptgbnet}& - & - & 80.5 \\
 PRANet~\cite{ptpranet}& - & - & 81.0 \\
\hline
 Transformer~\cite{pointbert}& 79.86 & 80.55 & 77.24 \\
 Transformer-OcCo~\cite{pointbert}& 84.85 & 85.54 & 78.79 \\
 Point-BERT~\cite{pointbert}& 87.43 & 88.12 & 83.07 \\
 {\bf Point-MAE}& {\bf 90.02}&{\bf 88.29} &{\bf 85.18} \\
\hline
\end{tabular}
\end{center}
\end{table}

We conduct experiments on three variants: OBJ-BG, OBJ-ONLY, and PB-T50-RS. Details are provided in supplementary materials. Note that no voting methods or data augmentation are used during testing. The results are presented in Table \ref{tb:2}. Our Point-MAE largely improves the baseline by 10.16\%, 7.74\%, and 7.94\% for three variants respectively. On the hardest variant PB-T50-RS, our model achieves 85.18\% accuracy, outperforming Point-BERT~\cite{pointbert} by 2.11\%. Though being pre-trained on clean objects, our Point-MAE generalizes well on real-world data, presenting a strong generalization capability. 

\vspace{-5mm}

\subsubsection{Object Classification on clean objects dataset}
We evaluate our pre-trained model on ModelNet40~\cite{modelnet} for object classification. ModelNet40 consists of 12,311 clean 3D CAD models, covering 40 object categories. We follow standard protocols to split ModelNet40 into 9843 instances for the training set and 2468 for the testing set. Standard random scaling and random translation are applied for data augmentation during training. For fair comparisons, we also use the standard voting method~\cite{ptrscnn} during testing. More details are provided in supplementary materials.

\begin{table}
\begin{center}
\vspace{-3mm}
\caption{{\bf Object classification on ModelNet40.} We compare our approach with various self-supervised (left) and supervised (right) methods. [T] represents the model is based on modified Transformers. [ST] represents the standard Transformers models.}

\label{tb:1}
\begin{minipage}{0.45\textwidth}
\begin{tabular}{clc}
\hline
&Self-supervised methods   & Accuracy\\
\hline
&OcCo~\cite{ptocco}  & 93.0\% \\
&STRL~\cite{ptstrl}   & 93.1\% \\
&IAE~\cite{ptiae} & 93.7\% \\
&[ST]Transformer-OcCo~\cite{pointbert} & 92.1\% \\
&[ST]Point-BERT~\cite{pointbert} & 93.2\% \\
&[ST]{\bf Point-MAE}& {\bf 93.8\%} \\
\hline

\end{tabular}
\end{minipage}
\begin{minipage}{0.45\textwidth}
\begin{tabular}{clc}
\hline
& Supervised methods &    Accuracy\\
\hline
 &PointNet~\cite{pointnet}&  89.2\% \\
 &PointNet\(++\)~\cite{pointnet++}&   90.7\% \\

 &PointCNN~\cite{ptptcnn}&  92.5\% \\
 &KPConv~\cite{ptkpconv}&  92.9\% \\
&DGCNN~\cite{ptptcnn}& 92.9\% \\
 &RS-CNN~\cite{ptrscnn}& 92.9\% \\
 &[T]PCT~\cite{ptpct} &  93.2\% \\
 &[T]PVT~\cite{ptpvt} &  93.6\% \\
 &[T]PointTransformer~\cite{ptpt}& 93.7\% \\
 &[ST]Transformer~\cite{pointbert}& 91.4\%\\
\hline
\end{tabular}
\end{minipage}
\end{center}
\end{table}


Experiment results are presented in Table \ref{tb:1}. For fair comparisons, all the reported methods are given 1024 points that only contain coordinate information without any normal information. Our Point-MAE achieves 93.8\% accuracy, improving 2.4\% accuracy compared to training from scratch (91.4\%). Compared with other self-supervised learning methods, our Point-MAE achieves state-of-the-art performance. Specifically, our approach with standard Transformers backbone surpasses IAE~\cite{ptiae} that uses a more powerful DGCNN~\cite{ptdgcnn} as the backbone (As shown in Table \ref{tb:1}, when training from scratch, DGCNN achieves 92.9\% accuracy, which is much higher). Besides, Point-MAE outperforms sophisticated Point-BERT~\cite{pointbert} by 0.6\% accuracy. Note that this improvement is significant as ModelNet40 is a relatively small dataset. Besides, our approach surpasses all the dedicated Transformers models from supervised learning. Furthermore, given 8192 points as input, our Point-MAE achieves 94.04\% accuracy. 

\subsubsection{Few-shot Learning}

We follow previous works~\cite{pointbert,fewshot,ptocco} to conduct few-shot learning experiments on ModelNet40~\cite{modelnet}, adopting \(n\)-way, \(m\)-shot setting, where \(n\) is the number of classes that randomly selected from the dataset and \(m\) is the number of objects randomly sampled for each class. We use the above-mentioned \(n \times m \) objects for training. During testing, we randomly sample 20 unseen objects from each of \(n\) classes for evaluation. 

The results with the setting of \(n \in \left\{ 5, 10 \right\}\) and \(m \in \left\{ 10, 20 \right\} \) are presented in Table \ref{tb:few}. Following standard protocol, we conduct 10 independent experiments for each setting and report mean accuracy with standard deviation. Our Point-MAE significantly advances state-of-the-art accuracies of four settings by 1.5\%-2.3\%, with smaller deviations. 

\begin{table}
\vspace{-3mm}
\begin{center}
\caption{{\bf Few-shot object classification on ModelNet40.} We conduct 10 independent experiments for each setting and report mean accuracy (\%) with standard deviation.}
\setlength\tabcolsep{2pt}
\label{tb:few}
\begin{tabular}{l|cccc}
\hline
Methods & 5-way,10-shot & 5-way,20-shot & 10-way,10-shot &10-way,20-shot  \\

\hline
 DGCNN-rand~\cite{ptocco} & 31.6 \(\pm\) 2.8 & 40.8 \(\pm\) 4.6 & 19.9 \(\pm\) 2.1 & 16.9 \(\pm\) 1.5\\
 DGCNN-OcCo~\cite{ptocco} & 90.6 \(\pm\) 2.8 & 92.5 \(\pm\) 1.9 & 82.9 \(\pm\) 1.3 & 86.5 \(\pm\) 2.2\\
 Transformer-rand~\cite{pointbert} & 87.8 \(\pm\) 5.2 & 93.3 \(\pm\) 4.3 & 84.6 \(\pm\) 5.5 & 89.4 \(\pm\) 6.3\\
 Transformer-OcCo~\cite{pointbert} & 94.0 \(\pm\) 3.6 & 95.9 \(\pm\) 2.3 & 89.4 \(\pm\) 5.1 & 92.4 \(\pm\) 4.6\\
 Point-BERT~\cite{pointbert} & 94.6 \(\pm\) 3.1 & 96.3 \(\pm\) 2.7 & 91.0 \(\pm\) 5.4 & 92.7 \(\pm\) 5.1\\
 {\bf Point-MAE} & {\bf 96.3 \(\pm\) 2.5}&{\bf 97.8 \(\pm\) 1.8} & {\bf 92.6 \(\pm\) 4.1} & {\bf 95.0 \(\pm\) 3.0}\\

\hline
\end{tabular}
\vspace{-10mm}
\end{center}

\end{table}

\vspace{-2mm}
\subsubsection{Part Segmentation}

We evaluate the representation learning capability of our Point-MAE on ShapeNetPart dataset~\cite{shapenetpart}, which contains 16,881 objects covering 16 categories. We follow previous works~\cite{pointnet,pointnet++,pointbert} to sample 2048 points as input for each object, which results in 128 point patches. Our segmentation head is relatively simple and does not use any propagating operation or DGCNN~\cite{ptdgcnn}. For fair comparisons, our segmentation head has a similar weight with Point-BERT~\cite{pointbert} and also uses learned features from \(4th\), \(8th\) and \(12th\) layer of Transformer block. We concatenate the three levels of features, then adopt average pooling and max pooling separately to obtain two global features. Besides, the concatenated features represent for 128 center points and are up-sampled~\cite{pointnet++} to 2048 input points to obtain features for each point. After concatenating per point features with two global features, MLP is adopted to predict the label for each point. More details are provided in supplementary materials. Note that no voting methods or data augmentation are used during testing.

As shown in Table~\ref{tb:seg}, we report mean IoU (mIoU) for all instances, with IoU for each category. Our Point-MAE achieves 86.1\% mIoU, improving the baseline by 1\% mIoU. Our Point-MAE with a simple segmentation head also outperforms Point-BERT~\cite{pointbert}, which uses DGCNN~\cite{ptdgcnn} and propagation in their segmentation head.

\begin{table}
\vspace{-2mm}
\begin{center}
\caption{{\bf Part segmentation on ShapeNetPart dataset.} We report mean IoU for all instances mIoU\(_I\) (\%), with IoU (\%) for each category. }

\label{tb:seg}
\begin{tabular}{lc|cccccccc}
\hline
Methods  & mIoU\(_I\) & aero & bag& cap& car& chair &e-phone&guitar &knife\\ 
 & &lamp &laptop &motor&mug&pistol&rocket&s-board&table  \\
\hline

 PointNet~\cite{pointnet} & 83.7 & 83.4 & 78.7 & 82.5 & 74.9 & 89.6 & 73.0 & 91.5 & 85.9 \\
                &      & 80.8 & 95.3 & 65.2 & 93.0 & 81.2 & 57.9 & 72.8 & 80.6 \\

 PointNet\(++\)~\cite{pointnet++} & 85.1 & 82.4 & 79.0 & 87.7 & 77.3 & 90.8 & 71.8 & 91.0 & 85.9 \\
                &      & 83.7 & 95.3 & 71.6 & 94.1 & 81.3 & 58.7 & 76.4 & 82.6 \\

 DGCNN~\cite{ptdgcnn} & 85.2 & 84.0 & 83.4 & 86.7 & 77.8 & 90.6 & 74.7 & 91.2 & 87.5 \\
                &      & 82.8 & 95.7 & 66.3 & 94.9 & 81.1 & 63.5 & 74.5 & 82.6 \\

\hline
 Transformer~\cite{pointbert} & 85.1 & 82.9 & 85.4 & 87.7 & 78.8 & 90.5 & 80.8 & 91.1 & 87.7 \\
                &      & 85.3 & 95.6 & 73.9 & 94.9 & 83.5 & 61.2 & 74.9 & 80.6 \\
 Point-BERT~\cite{pointbert} & 85.6 & 84.3 & 84.8 & 88.0 & 79.8 & 91.0 & 81.7 & 91.6 & 87.9 \\
                &      & 85.2 & 95.6 & 75.6 & 94.7 & 84.3 & 63.4 & 76.3 & 81.5 \\
 {\bf Point-MAE} & {\bf 86.1} & 84.3 & 85.0 & 88.3 & 80.5 & 91.3 & 78.5 & 92.1 & 87.4 \\
                 &      & 86.1 & 96.1 & 75.2 & 94.6 & 84.7 & 63.5 & 77.1 & 82.4 \\
\hline
\end{tabular}
\end{center}
\vspace{-8mm}
\end{table}

\subsection{Ablation Study}
\label{sec:mask}
\begin{table}
\vspace{-3mm}
\begin{center}
\caption{{\bf Ablation study on masking strategy.} We conduct experiments using two masking strategy with different masking ratios (\%), and report pre-train loss ($\times$ 1000) as well as fine-tune accuracy (\%).}
\setlength\tabcolsep{2pt}
\label{tb:mask}
\begin{tabular}{|lccc|lccc|lccc|}
\hline
Type & Ratio & Loss & Acc. \quad &Type & Ratio & Loss & Acc. \quad &Type & Ratio & Loss & Acc. \quad  \\
\hline
Block & 40 & 2.83 &  92.67 \quad &Random & 40 & 2.49 & 92.46 \quad &Random & 70 & 2.68 & 93.11 \quad\\
Block & 60 & 2.89 & 92.67 \quad &Random & 50 & 2.54 & 92.43 \quad &Random & 80 & 2.77 & 93.03 \quad \\
Block & 80 & 2.98 & 92.50 \quad &Random & 60 & 2.60 & {\bf 93.19} \quad &Random & 90 & 2.89 & 92.63 \quad \\
\hline
\end{tabular}
\end{center}
\vspace{-13mm}
\end{table}

\subsubsection{Masking Strategy}
\begin{figure}
    \centering
    \includegraphics[width=\textwidth]{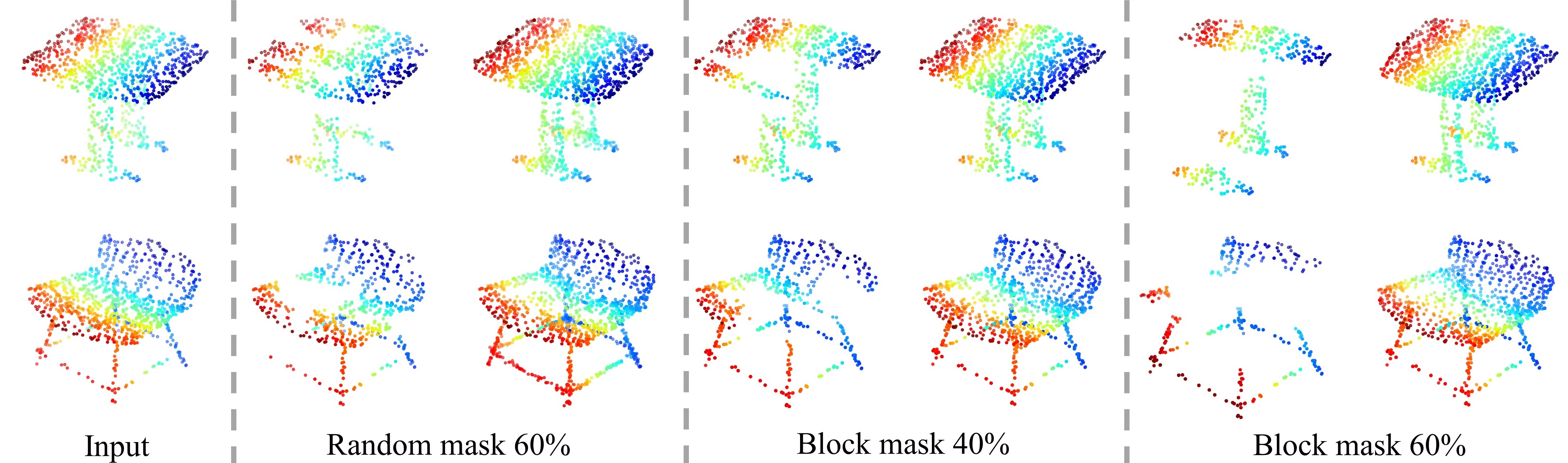}
    \caption{{\bf Reconstructions with different masking strategies.} We mainly show three different masking strategies for same inputs (leftmost). In each column, masked inputs (left) and reconstructions (right) are shown. Instances are from ShapeNet validation set.}
    \label{fig:mask}
\end{figure}

To find a proper masking strategy for our method, we compare two masking types with different masking ratios. No voting method is used during testing. The reconstruction loss and fine-tune accuracy on ModelNet40 are presented in Table \ref{tb:mask}. We also visualize reconstructions with different masking strategies in Figure \ref{fig:mask}. 

The block masking~\cite{pointbert,cv2} type masks neighbouring point patches, resulting in masked blocks. Though this strategy is harder for reconstruction, adopting a medium masking ratio can also achieve good performance.

The random masking type masks random point patches and empirically results in the best performance with a high masking ratio (i.e. 60\%-80\%). The performance degrades largely with low making ratios and also degrades slightly if the masking ratio is too high. 

\subsubsection{Effect of shifting mask tokens}

Our Point-MAE shifts mask tokens from the input of the encoder to the lightweight decoder. To demonstrate the effectiveness of this design, we conduct an experiment in which the mask tokens are processed from the input of the encoder. For fair comparisons, the autoencoder's backbone adopts the same encoder and prediction head as Point-MAE but without the decoder, resulting in the exact same model on fine-tune tasks. We use random masking at a ratio of 60\% in this experiment. After pre-training, a smaller reconstruction loss is observed (2.51), compared to Point-MAE (2.60). For the fine-tune performance on ModelNet40, it achieves 92.14\% accuracy, much lower than Point-MAE (93.19\%). This result is not surprising and can be explained. At the input of the encoder, all the tokens, including mask tokens, must be provided with location information by positional embeddings. This causes early leakage of location information because mask tokens are processed for the reconstruction of point patches in coordinate space. The leakage of location information makes the reconstruction task less challenging, and the model cannot learn latent features well, leading to worse fine-tune performance.

\section{Conclusions}

In this paper, we present a novel scheme of masked autoencoders for point cloud self-supervised learning, termed as Point-MAE. Our Point-MAE is neat and efficient, with minimal modifications based on the properties of the point cloud. The effectiveness and high generalization capability of our approach are verified on various tasks, including object classification, few-shot learning, and part segmentation. Specifically, Point-MAE outperforms all the other self-supervised learning methods. We also show with our approach, a simple architecture that is entirely based on standard Transformers can surpass dedicated Transformer models from supervised learning. Furthermore, our work inspires the feasibility of applying unified architectures from languages and images to the point cloud.

%
%
\bibliographystyle{splncs04}
\bibliography{egbib}
\end{document}